\documentclass[letterpaper, 10 pt, conference]{ieeeconf}  

\IEEEoverridecommandlockouts                              

\usepackage{graphicx} 
\usepackage{amsmath} 
\usepackage{amssymb}  
\usepackage{cite}
\usepackage{hyperref} 
\usepackage{subcaption}
\usepackage{hyperref}
\usepackage{caption}
\usepackage{float}

\newcommand{\comment}[1]{}
\usepackage{color}

\title{\LARGE \bf
Morphing MILR: Design and control of a cable-driven limbless robot with rolling joints for maneuvering in complex environments
}

\author{
Anonymous authors
}

\author{%
Donoven Dortilus$^{\dagger}$, Tianyu Wang$^{\dagger}$, Galen Tunnicliffe, Matthew Fernandez, Daniel I. Goldman%
\thanks{$\dagger$These authors contributed equally to this work.}
\thanks{Donoven Dortilus, Tianyu Wang, Galen Tunnicliffe, Matthew Fernandez, and Daniel I. Goldman are with Georgia Institute of Technology, Atlanta, GA 30332, USA. {\tt\small \{ ddortilus3, tianyuwang, gtunnicliffe3, mfernandez64\}@gatech.edu, daniel.goldman@physics.gatech.edu}}%
}

\begin{document}
\maketitle
\thispagestyle{empty}
\pagestyle{empty}

\begin{abstract}

Limbless robots offer exceptional mobility in confined and cluttered environments due to their slender bodies and their ability to exploit body-terrain interactions. Recent designs incorporating compliance demonstrate robust locomotion without complex sensing or control; however, these systems typically rely on fixed body configurations, with each morphology specialized for a single locomotion mode or environment. This raises a key challenge: how can a single limbless robot achieve versatile locomotion while preserving the robustness of compliance-mediated locomotion? To address this challenge, we present a cable-driven limbless robot that reconfigures body morphology and compliance to enable diverse locomotion modes. Distributed cable actuation generates traveling body waves, while programmable passive compliance enables robust contact-rich locomotion without terrain knowledge or high-bandwidth feedback. Rolling joints reorient bending planes along the body, enabling rapid reconfiguration and smooth transitions between locomotion styles, and incorporate geared locking to maintain configuration without continuous power. By combining programmable bending compliance and morphology control, the platform achieves lateral undulation, sidewinding, rolling, and twisting within a single system. Experiments demonstrate reliable gait generation, traversal in obstacle-rich environments, and transitions between modes, establishing a versatile limbless platform for navigating complex environments with applications in search and rescue, environmental monitoring, and inspection.

\end{abstract}

\section{Introduction}\label{sec:intro}

Limbless robots show strong potential for maneuvering through complex terrain for applications such as search and rescue, environmental monitoring, and industrial inspection, largely because their slender, hyper-redundant bodies can access spaces that conventional mobile robots cannot \cite{hirose1993biologically,hirose2004biologically,liljeback2012review,wright2007design,wright2012unified,rollinson2014series}. A wide range of engineered mechanisms and bio-inspired control strategies replicate limbless locomotion and synthesize gaits in open and obstacle-rich settings \cite{crespi2008onlineopt,crespi2005amphibotI,hirose2004biologically,transeth2008snake,tesch2009parameterized,travers2016shapebased,saito2002serpentine,gong2016kinematic,kano2013scaffold,sanfilippo2017perception}. Many platforms can achieve out-of-plane motion, including sidewinding, lifting, climbing, and rolling \cite{burdick1993sidewinding,marvi2014sidewinding,astley2015modulation,zhen2015rolling,takemori2018ladder,takemori2021hooppassing,takemori2022adaptive,wang2021reconstruction}, but these capabilities come from increased sensing, higher-bandwidth feedback, and carefully tuned control that can become uncertain when contact conditions change or when the environment is only partially known \cite{sanfilippo2017perception,hanssen2020path,sartoretti2021autonomous,ruscelli2018proprioceptive,tanaka2015rangesensor,bing2020perceptionaction,ramesh2022sensnake}.

\begin{figure}[t]
\centering
\includegraphics[width=1\columnwidth]{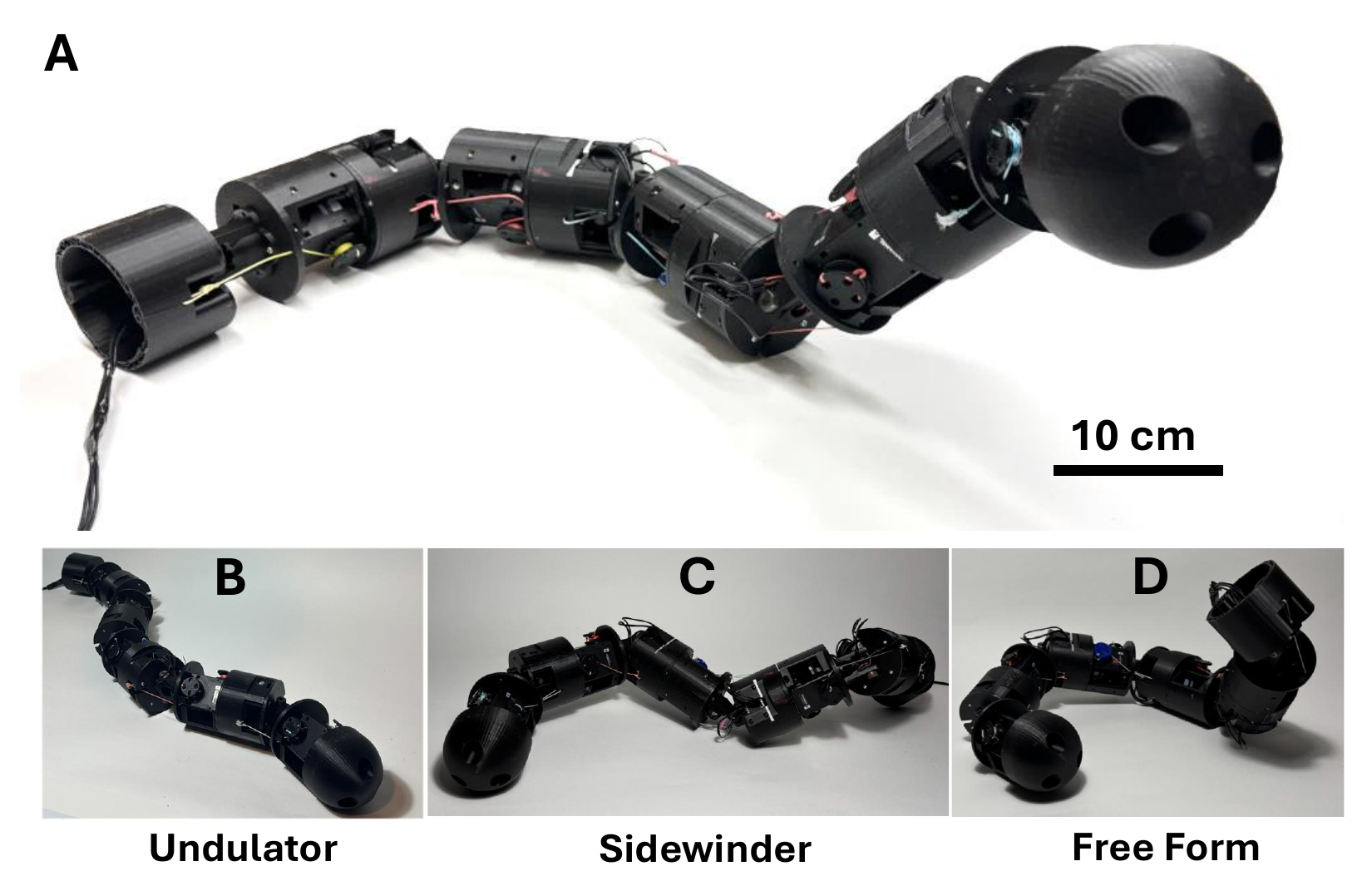}
\caption{(A) The limbless robot Morphing MILR designed for locomotion in complex environments, capable for quick morphology switching, such as planar undulator (B), sidewinder (C), and many more flexible morphologies (D).}
\label{fig:intro_robot}
\end{figure}

In parallel, a smaller set of limbless robots demonstrates an alternative strategy that embeds \emph{mechanical intelligence} through morphology, passive compliance, and contact-mediated interactions, allowing body-environment dynamics to contribute directly to effective locomotion \cite{sitti2021physical,fu2020robotic,travers2016shape,serrano2015incorporating}. Within this paradigm, cable-driven bilateral actuation with programmable compliance simplifies control in heterogeneous terrestrial environments by enabling robust obstacle-aided undulation without explicit terrain knowledge \cite{wang2023mechanical}. More recently, these design principles extended to additional gait families and domains, including sidewinding facilitated by anisotropic compliance \cite{kojouharov2024anisotropic} and undulatory swimming in complex aquatic terrain \cite{wang2025aquamilr,fernandez2025aquamilrplus}. However, such systems remain largely tied to fixed hardware morphologies and, in several cases, a fixed dominant bending plane or gait-specific configuration, limiting the diversity of locomotion styles that a single device can express.

In this work, we extend compliance-enabled cable-driven undulation beyond a single configuration by introducing Morphing MILR (Mechanically Intelligent Limbless Robot) (Fig.~\ref{fig:intro_robot}), a modular platform that rapidly reconfigures its morphology to realize multiple limbless locomotion modes within a three-dimensional workspace, a capability not present in prior mechanically intelligent works \cite{wang2023mechanical,kojouharov2024anisotropic}. The robot consists of repeated modules that combine cable-driven bending with rolling joints that reorient the bending planes. By actively controlling where and how we express changes in angles, Morphing MILR unifies previously separate cable-driven limbless robot morphologies within a single architecture, enabling transitions between modes such as lateral undulation, sidewinding, and rolling while preserving the compliance that facilitates robust body-environment interaction. Through experiments, we demonstrate distinct movement styles and reliable mode transitions driven primarily by morphology and compliance rather than terrain-specific sensing. These capabilities open pathways toward field-relevant behaviors in harsh, confined, and contact-rich environments.


\section{Robot Design}
\label{sec:design}

\begin{figure}[t]
\centering
\includegraphics[width=1\columnwidth]{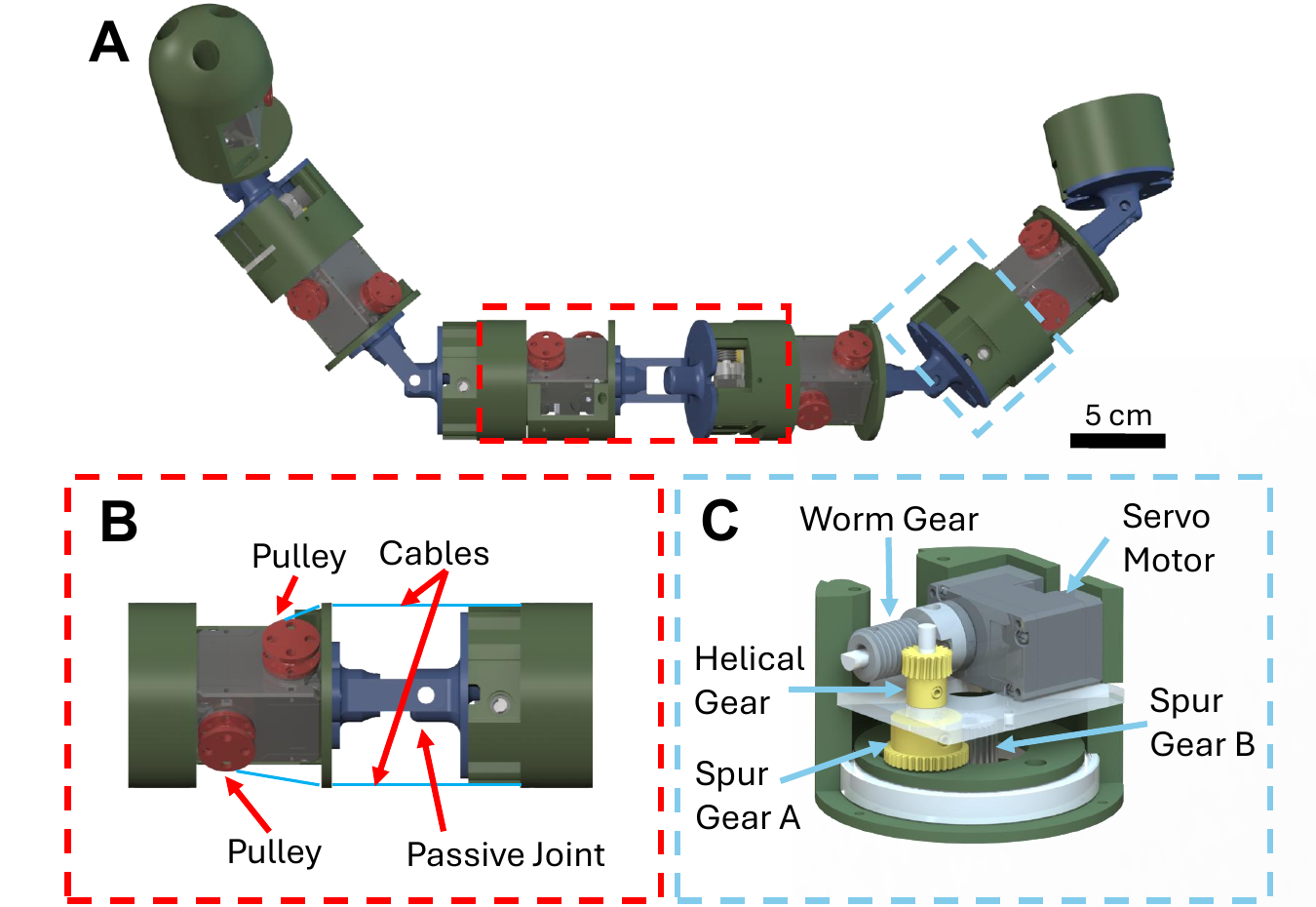}
\caption{Overview of full design (A) broken down into bilateral cable actuation between modules (B) with rotating geared down servo motor controlling module rotation (C).}
\label{fig:robot_design}
\end{figure}

\subsection{Design Overview}

Morphing MILR currently consists of six repeated modules, with the potential to add more (Fig.~\ref{fig:robot_design}A), each integrating a bending joint actuated by antagonistic cables and a rolling base joint that reorients the bending plane. This architecture preserves the mechanically intelligent behavior exhibited by the bilateral actuation mechanism \cite{wang2023mechanical}, while no longer being confined to a single combination of joint orientation, allowing for a wider range of traversable terrains in a single architecture and enabling fast transitions between locomotion modes such as lateral undulation, sidewinding, and rolling. 

\begin{table}[h]
\centering
\renewcommand{\arraystretch}{1.5} 
\begin{tabular}{l l} 
\hline\hline 
\\[-1.25em] 
\textbf{Mass} & \begin{tabular}[c]{@{}l@{}}Single Module: 0.25 kg \\ Full 6 Module Robot: 1.5 kg \end{tabular} \\ 
\hline
\textbf{Dimensions} & \begin{tabular}[c]{@{}l@{}}6.5 cm Diameter\\ 70 cm Length\end{tabular} \\ 
\hline
\textbf{Power} & 12 V, 1 A (normal operation) \\ 
\hline
\textbf{Communication} & \begin{tabular}[c]{@{}l@{}} RS-485 Serial (internal)\end{tabular} \\ 
\hline
\textbf{Sensing} & \begin{tabular}[c]{@{}l@{}}Individual cable length and tension\\ Relative module orientation\end{tabular} \\ 
\hline
\textbf{Actuation} & \begin{tabular}[c]{@{}l@{}}Cable Motors: 1.4 Nm (stall)\\ Rolling Base Motors: 0.92 Nm (stall)\end{tabular} \\[-1.25em] \\ 
\hline\hline 
\\[-1.25em] 
\end{tabular}
\caption{Morphing MILR specifications.}
\end{table}

\begin{figure}[t]
\centering
\includegraphics[width=1\columnwidth]{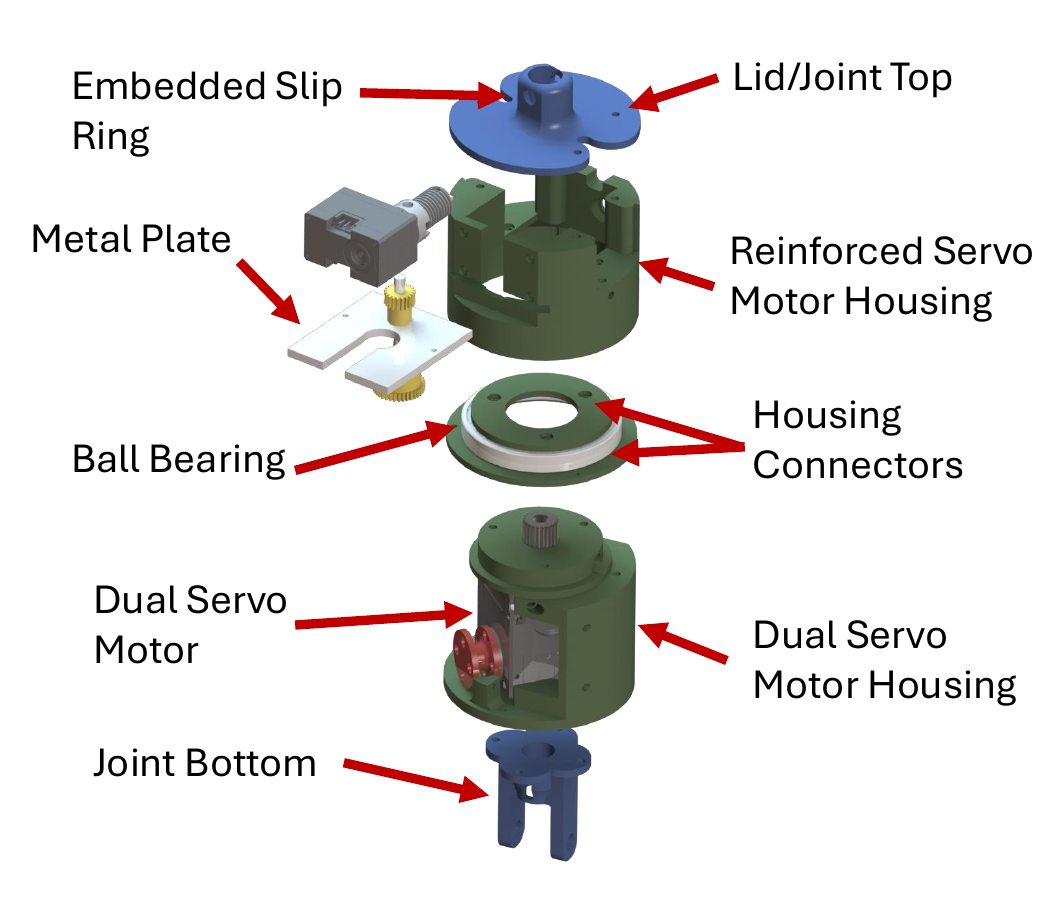}
\caption{Exploded view of the repeated module highlighting major structural components.}
\label{fig:exploded}
\end{figure}

\subsection{Rolling Base Design}

The rolling base must reorient partially supported body segments, overcome interface friction, and in demanding configurations, lift downstream modules during locomotion transitions (Fig.~\ref{fig:robot_design}C). These requirements motivated a compact, high torque geared transmission constrained within a reinforced housing. An aluminum plate integrated into the housing provides the necessary reinforcement that mitigates the flexing of the printed part. This keeps gear engagement consistent while under heavier loads. A compact servo motor, Robotis XC330,  applies torque through the worm, helical, and spur stages to drive the rotating base, with all gears mounted on bearing-constrained shafts. We determined the necessary overall gear reduction based on the maximum moment required to lift two adjacent modules in the most demanding joint orientation. The gear ratio of the current packaging is 1:13, including a standard safety factor. A set screw secures spur gear B to the housing of the dual-axis servo motor. The dual servo housing sits within the inner race of the main ball bearing and the reinforced housing sits around the outer race. Two connection plates constrain the housings axially while still allowing free spin (Fig.~\ref{fig:exploded}). The worm gear within the transmission assembly makes the system non-backdrivable, allowing the robot to maintain configuration without continuous motor power. 
 
\subsection{Bilateral Cable Actuation}

\begin{figure}[t]
\centering
\includegraphics[width= 1\columnwidth]{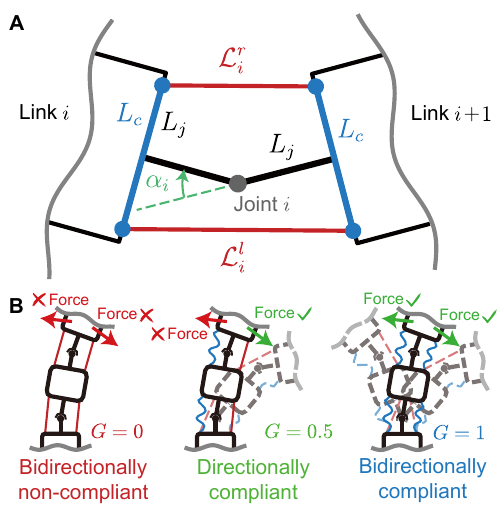}
\caption{Programmable body compliance through bilateral cable actuation mechanism. (A) A geometric model illustrating a single joint (B) A schematic displaying various compliance states based on the generalized compliance variable $G$. Figure adapted from~\cite{wang2023mechanical}.}
\label{fig:G explain}
\end{figure}

Each bending joint uses a bilateral pulley-cable actuation scheme that models musculoskeletal actuation in limbless locomotion (Fig. \ref{fig:robot_design}B). A dual-axis servo motor, Robotis 2XL430, drives opposing pulleys that lengthen and shorten high-strength, inextensible, braided line to set the joint angle.

The module structure incorporates the cable routing path. The bending joint between adjacent modules has a shoulder bolt that holds it together. The cable pair is on a plane perpendicular to the joint axis, so different lengths cause a bending angle for the joint up to 70 degrees, limited by module self collision. Coordinated cable actuation enables bending joint angle control and body-wave propagation. This bilateral actuation scheme also enables modulation of joint compliance through cable length control, supporting adaptive terrain interaction. 
 
 \subsection{Electrical}
All actuators are daisy chained on a common power and communication bus, minimizing wiring while enabling coordinated control across the entire body. We used an external motor controller (Robotis U2D2) to communicate with the actuator network over a serial link, allowing command of the bending cables and rolling base independently or in coordinated groups, which is necessary to achieve the mechanical compliance. We placed slip rings, with wires rated to handle a current of 3 A, between modules  to connect the shared bus across joints, enabling continuous rolling reconfiguration while preserving uninterrupted power and communication without twisting internal wiring. Currently, we operate while tethered to a power supply and an external computer running a script to change between gaits. 

\begin{figure}[t]
\centering
\includegraphics[width= .9\columnwidth]{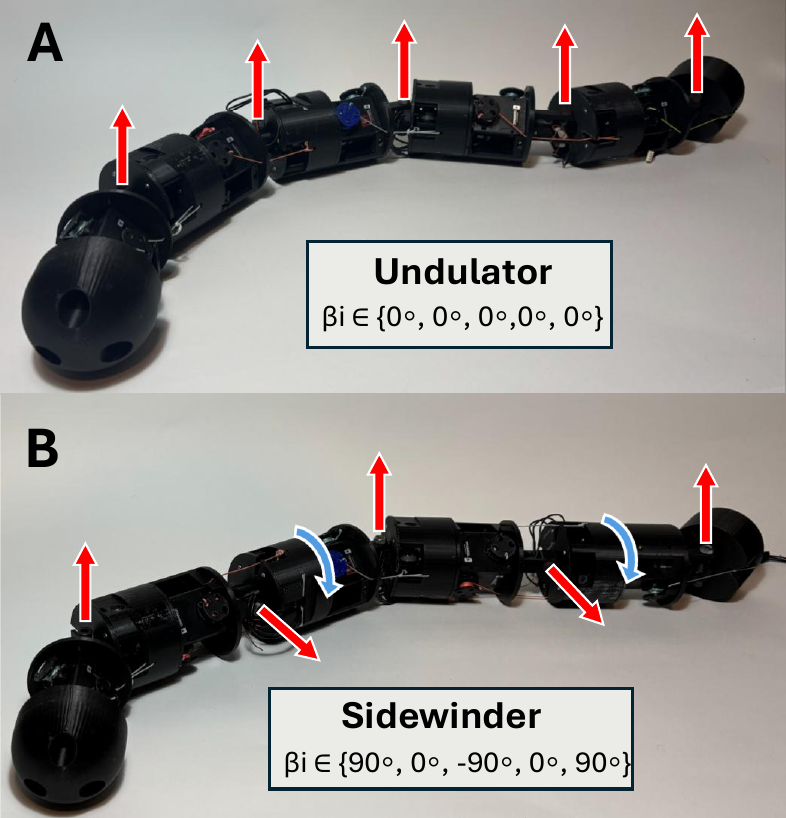}
\caption{Transitioning from lateral undulation to sidewinding via simultaneous rolling joint movement.}
\label{fig:robot_range of motion}
\end{figure}

\section{Robot Control}

\subsection{Joint Control with Bilateral Actuation}

To realize a joint angle $\alpha_i(t)$ on the $i$-th joint, the left and right cable lengths around each bending joint must be set appropriately. The required cable lengths primarily come from the robot geometry and the joint rotation range \cite{wang2023mechanical}. Let $\mathcal{L}^l$ and $\mathcal{L}^r$ denote the left and right cable lengths across joint $i$ \cite{wang2023mechanical}. The equations for cable lengths are:
\begin{equation}
\begin{aligned}
    \mathcal{L}^l(\alpha) &= 2\sqrt{L_{c}^2 + L_{j}^2} \cos\left[-\frac{\alpha}{2}+\tan^{-1}\left(\frac{L_{c}}{L_{j}}\right)\right]\\
    \mathcal{L}^r(\alpha) &= 2\sqrt{L_{c}^2 + L_{j}^2} \cos\left[\frac{\alpha}{2}+\tan^{-1}\left(\frac{L_{c}}{L_{j}}\right)\right]
\end{aligned}
\label{eq:ExactLength}
\end{equation}

Bilateral cables also enable body compliance, as outlined in \cite{wang2023mechanical}. A generalized compliance variable (G) alters the robot's body compliance using these equations:

\begin{equation}
\begin{array}{l}
L_{i}^l(\alpha_{i}) = \left\{\begin{array}{llc}{\mathcal{L}_{i}^l(\alpha_{i}),} & \,{\text{if } \alpha_{i} \leq -\gamma} \\ {\mathcal{L}_{i}^l[-\min(A, \gamma)]+l_0\cdot[\gamma + \alpha_{i}],} & \,{\text{if } \alpha_{i} > -\gamma}\end{array}\right. \\ 
L_{i}^r(\alpha_{i}) = \left\{\begin{array}{llc}{\mathcal{L}_{i}^r(\alpha_{i}),} &\ \ \, {\text{if } \alpha_{i} \geq \gamma} \\{\mathcal{L}_{i}^r[\min(A, \gamma)]+l_0\cdot[\gamma - \alpha_{i}],} & \ \ \, {\text{if } \alpha_{i} < \gamma}\end{array}\right.
\end{array}
\label{eq:policy}
\end{equation}

where $\gamma = (2G_i - 1)A$ and $l_0$ is a design parameter that determines the amount of cable slack introduced. Detailed derivation and parameter selection comes from \cite{wang2023mechanical}. In this work, we only keep compliance active during obstacle interaction and gait changes, with $G_i = 1$ applied uniformly across modules. In the following sections, we present high-level shape-based gait templates; layering compliance modulation onto these templates when terrain interaction requires additional adaptability.

The passive body compliance enables the use of open-loop joint commands while still remaining effective across varied terrain as displayed in past works \cite{wang2023mechanical,wang2025aquamilr,kojouharov2024anisotropic,fernandez2025aquamilrplus,wang2026three}. The rolling-base morphology primarily allows transitions between the gait configurations, not requiring closed-loop control since the joints are mainly static during normal operation.

\subsection{Gait Implementation}
\subsubsection{Lateral Undulation}
Lateral undulation is a purely 2D morphology, aligning all rolling bases on a common plane (Fig.~\ref{fig:robot_range of motion}A). We denote the rotation axis of the body bending joint as $\beta_i$ in module $i$, 
 \begin{equation}
\beta_i \in \left\{0^\circ,\ 0^\circ,\ 0^\circ,\ 0^\circ,\ 0^\circ\right\}.
\label{eq:lateral_beta}
\end{equation}

\begin{figure}[t]
\centering
\includegraphics[width=1\columnwidth]{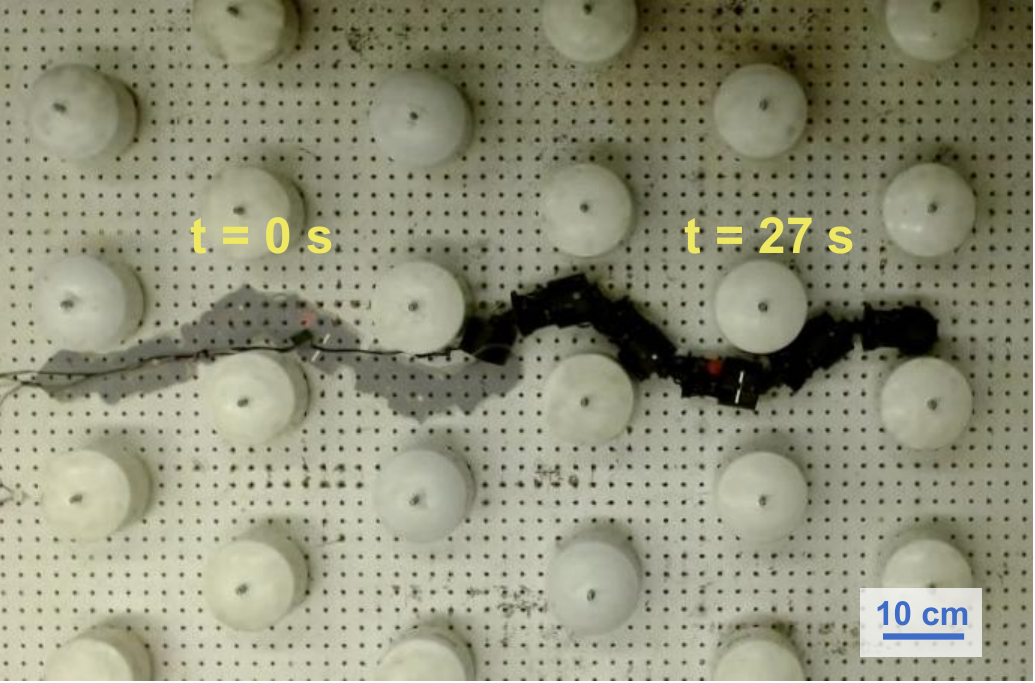}
\caption{Lateral undulation, leveraging cluttered terrain to achieve motion, 20 cycles at $\omega$ = 2 Hz.} 
\label{fig:Lateral undulation}

\end{figure}

Note, that $\beta_i$ is in reference to the previous module and is calculated based on the encoder values with in servos. Lateral undulation is produced via a head-to-tail propagating body wave (Fig.~\ref{fig:Lateral undulation}) according to the ``serpenoid" gait template \cite{hirose1993biologically}. The $i$-th bending joint angle, $\alpha_i$, is commanded as
\begin{equation}
\alpha_i(t) = A\sin\left(2\pi\xi \frac{i}{N} - 2\pi\omega t\right)
\label{eq:wave}
\end{equation}
where $A$ is the wave amplitude, $\xi$ is the spatial frequency which is number of periods expressed on the body, $\omega$ is the temporal frequency being the propagation speed of the wave over the body, $i$ is the joint index, and $N$ is the total number of bending joints. This wave is then propagated along the body for a set amount of cycles, where one cycle corresponds to a full wave traveling the full length of the robot.

\vspace{.8em}
\subsubsection{Sidewinding}
Sidewinding can be modeled as the coordinated superposition of orthogonal body waves \cite{astley2015modulation}, with limbless robots implementing sidewinding through shape-based gaits and control strategies \cite{burdick1994sidewinding,chong2021frequency,kojouharov2024anisotropic}. In Morphing MILR, the synthesizing of the required orthogonal components is done morphologically by fixing each rolling base to a prescribed orientation pattern that alternates the bending plane along the body (Fig.~\ref{fig:robot_range of motion}B). We assign
\begin{equation}
\beta_i \in \left\{90^\circ,\ 0^\circ,\ -90^\circ,\ 0^\circ,\ 90^\circ\right\} \quad
\label{eq:sidewinding_beta}
\end{equation}
which divides the robot into alternating, orthogonal bending planes. In this configuration we describe the joint angles as:
\begin{equation}
\alpha_i(t)=
\begin{cases}
A_H \sin\!\left(2\pi\xi \frac{i}{N} - 2\pi\omega t\right), & i \ \text{odd},\\[6pt]
A_V \sin\!\left(2\pi\xi \frac{i}{N} - 2\pi\omega t - \frac{\pi}{2}\right), & i \ \text{even}.
\end{cases}
\label{eq:rolling}
\end{equation}
$A_H$ and $A_V$ are the curvature amplitudes for the horizontal and vertical waves with the $A_H$ kept lower than $A_V$. This produces the characteristic sidewinding body waves, with alternating lifted and contacting segments that generate traveling contact patches and net translation with reduced slip \cite{marvi2014sidewinding,kojouharov2024anisotropic} (Fig.~\ref{fig:Sidewinding}).

\subsubsection{Rolling}
\vspace{.8em}
Rolling uses the same morphology as sidewinding, having the rolling bases remain fixed to the repeating orientation sequence in Eq.~\eqref{eq:sidewinding_beta} and utilizes similar equations for describing body waves as sidewinding being:
\begin{equation}
\alpha_i(t)=
\begin{cases}
A \sin\!\left(2\pi\xi \frac{i}{N} - 2\pi\omega t\right), & i \ \text{odd},\\[6pt]
A \sin\!\left(2\pi\xi \frac{i}{N} - 2\pi\omega t - \frac{\pi}{2}\right), & i \ \text{even}.
\end{cases}
\label{eq:rolling_alpha}
\end{equation}
Here $A$, the curvature amplitude, is the same for both waves and $\xi$, the spatial frequency, is kept below one. This results in perpendicular motion similar to that of sidewinding with a lower profile (Fig.~\ref{fig:Rolling}).

\subsubsection{Twisting}
In the twisting configuration, all cable-driven bending joints maintain a straight posture,
\begin{equation}
\alpha_i(t) = 0 \quad \forall i
\label{eq:twisting_alpha}
\end{equation}
generating motion using only the rolling base joints (Fig.~\ref{fig:Twisting}). Let $\beta_i(t)$ denote the current rolling-base angular position of module $i$. We actuate the rolling joints with constant, slightly offset angular velocities,
\begin{equation}
\beta_i(t) = \beta_i(0) + \omega t, 
\quad \Omega_i = \omega\hat{e}_i + \Omega_{(i-1)}, \quad \omega \ \text{given}
\label{eq:twisitng_beta}
\end{equation}
where $\omega$ is a constant  angular velocity commanded at every rolling joint. $\Omega_i$ is the $i^{th}$ module's angular velocity vector w.r.t. the inertial frame, with $\hat{e}_i$ as the local joint axis.  This generates a velocity gradient w.r.t the inertial frame and produces a distributed twist along the body, allowing the robot to rotate in place on flat ground.

\begin{figure}[t]
\centering
\includegraphics[width=1\columnwidth]{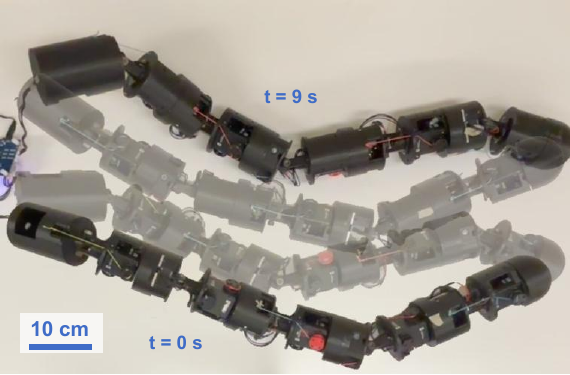}
\caption{Sidewinding, orthogonal serpenoid waves at differing amplitudes leading to motion perpendicular to orientation, 10 cycles at $\omega$ = 4 Hz.}
\label{fig:Sidewinding}
\end{figure}

\subsection{Morphology Control for Gait Transition}

Morphology transitions are executed by commanding a set of target rolling angles across the module chain, effectively remapping the direction in which each joint bends. During the transition, we return the robot to a home state with aligned joint planes and straight joints , then we command the rolling bases to the next configuration before undulation resumes. This approach maximizes repeatability and minimizes geometric coupling between gait patterns.

We can use programmable compliance to improve robustness. By temporarily increasing compliance in the body, the robot can accommodate incidental contacts and geometric constraints, while rolling bases reorient the bending planes \cite{wang2023mechanical,travers2016shape}. This compliance-mediated transition helps prevent jamming in tight spaces, reduces peak transmission loads, and allows the body to settle into the new gait before full-amplitude locomotion resumes. Together, rolling base reorientation and compliance modulation provide a compact framework for rapidly switching between gaits and for expanding to future motion primitives beyond rolling, sidewinding, and lateral undulation.

\section{Evaluation}

\begin{figure}[t]
\centering
\includegraphics[width=1\columnwidth]{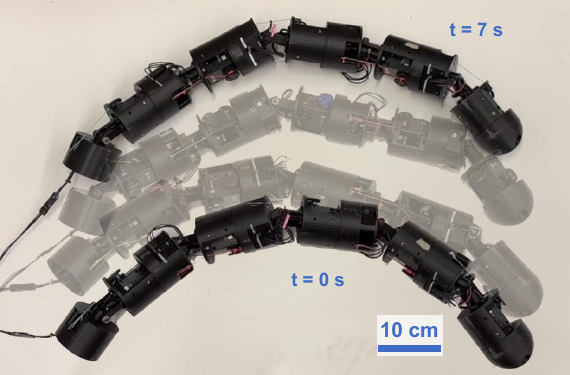}
\caption{Rolling, orthogonal serpenoid waves at at the same amplitude leading to motion perpendicular to orientation, 10 cycles at $\omega$ = 4 Hz.}
\label{fig:Rolling}
\end{figure}

\subsection{Obstacle Free Testing}

\begin{figure}[t]
\centering
\includegraphics[width=1\columnwidth]{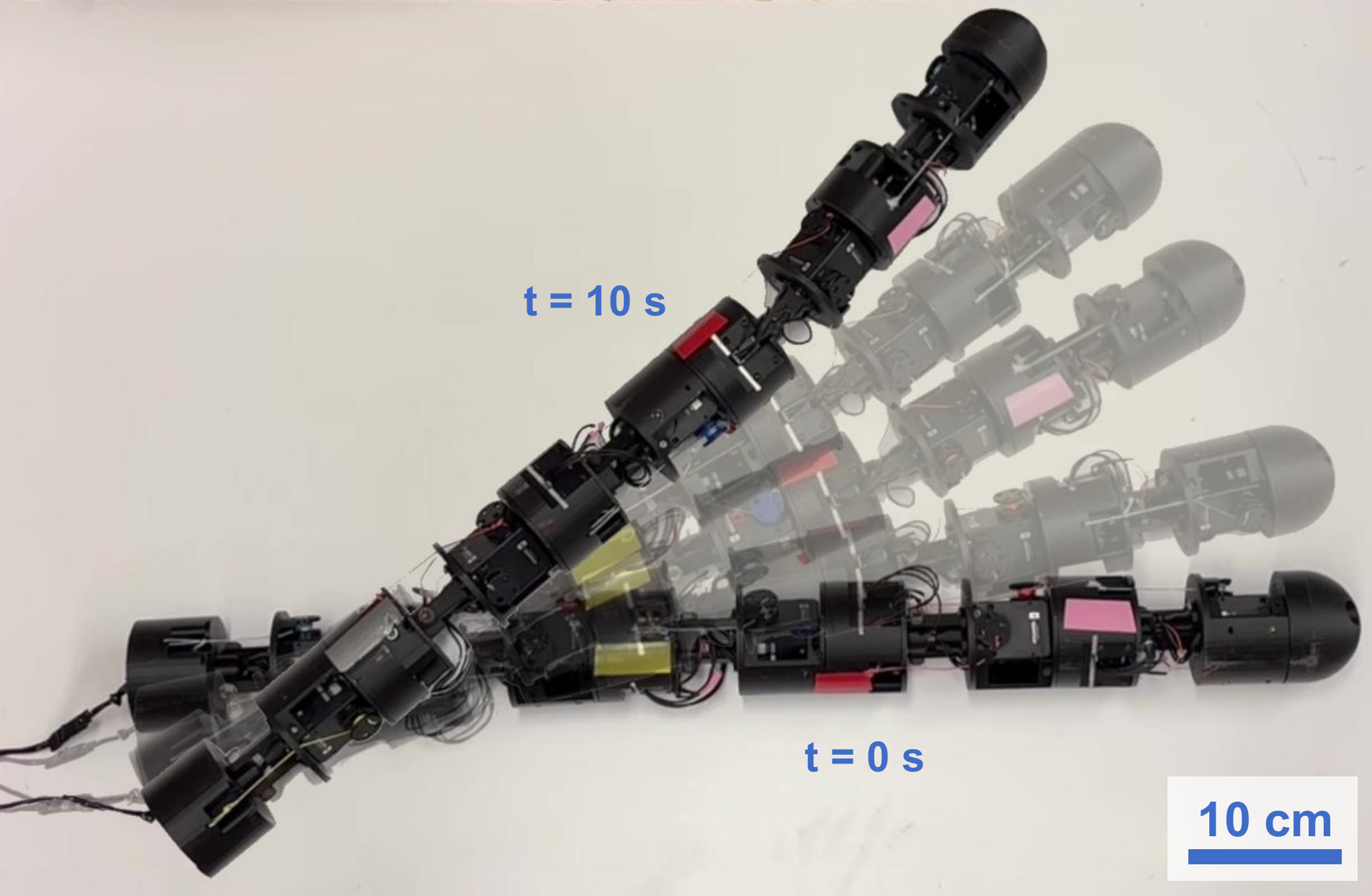}
\caption{Twisting, rotating modules at different angular velocities to turn in place, 10 cycles at $\omega$ = 4 Hz.}
\label{fig:Twisting}
\end{figure}

We tested whether Morphing MILR can change morphology reliably, ensuring that rolling bases can repeatedly reorient to a desired configuration. Generalized compliance was not used ($G=0$) in any of the obstacle free tests. In the 5 initial trials, the robot changes the alignments of the rolling bases to form the orthogonal wave pattern used in sidewinding and then runs 10 cycles. The sidewinding parameters utilized were $A_V = 30^\circ$, $A_H = 60^\circ$ with a spatial frequency of 1.2 and a temporal frequency of 4 Hz (Fig.~\ref{fig:Sidewinding}). The next 5 trials conducted were for rolling and used an $A = 60^\circ$ for both waves, with a spatial frequency of 1.2 and a temporal frequency of 4 Hz (Fig.~\ref{fig:Rolling}). The third set of 5 trials were done in twisting mode, with individual module rotation speeds varying from $5.6$ to $22.5$ $^\circ/s$ to see if any issues arose when more than a full rotation was performed (Fig.~\ref{fig:Twisting}). 

Across these trials, we use a pass fail checklist for the different key parameters. Wave amplitudes and rolling base angles must be within $\pm 5^\circ$ of the commanded value accounting for gear backlash and cable re-spooling differences. The spatial and temporal frequencies should match the instructed values with no drifting over cycles, verified over video footage. For the last 5 trials, the main criterion is the measured individual module speeds matching coded values in captured videos. In the trials performed, the robot passed all above criteria checks. 

\subsection{Lattice Testing}

Lattice experiments tested whether Morphing MILR preserves mechanical intelligence in organized obstacle fields, specifically, that the robot can progress through a lattice without localized jamming, defined as the interruption of wave propagation from sharp localized loads \cite{wang2023mechanical,travers2016shape,schiebel2020robophysical}.

Varying G values over 20 trials, Morphing MILR achieved repeatable in-lattice progression consistent with previous observations that compliance and directional interaction effects can improve obstacle-aided undulation. With G being the focus of these trials, all other parameters were kept constant with $A = 60^\circ$, a spatial frequency of 1.1 and a temporal frequency of 2 Hz. Trial success is defined by the robot traveling to the end of the lattice with its head reaching open space. The majority of successful runs occurred with $G$ = 1 (Fig.~\ref{fig:Lateral undulation}) with lower values of $G$ more frequently leading to localized jamming and higher values leading to a lack of forward progression \cite{wang2023mechanical,wang2020directional}. Overall, Morphing MILR successfully preserved the mechanically intelligent benefits of bilateral cable actuation.

\begin{figure*}[t]
\centering
\includegraphics[width=1\textwidth]{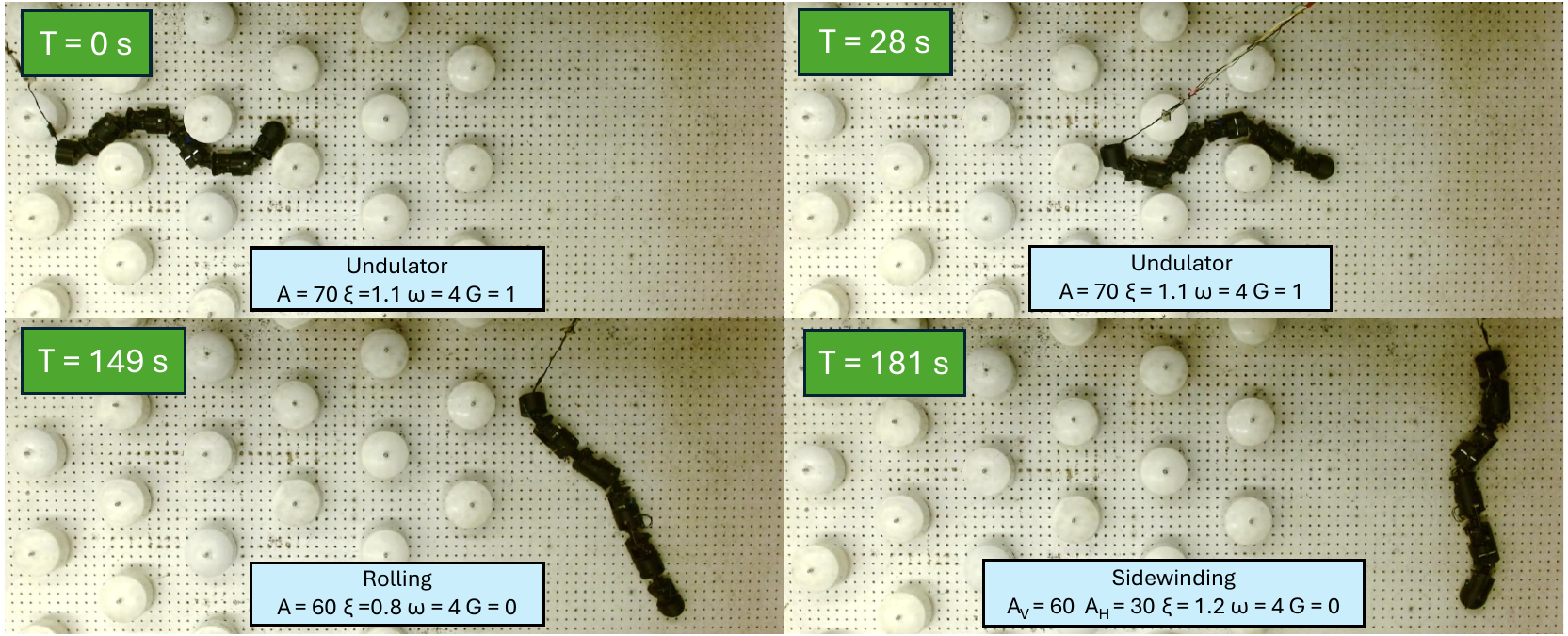}
\caption{Morphing MILR transitions from obstacle rich lattice locomotion via compliant lateral undulation to open terrain sidewinding, using rolling gait to transition environments.}
\label{fig:transition}
\end{figure*}

\subsection{Environment Transition Testing}

We designed environment transition trials to test the robot’s ability to maintain locomotion while moving between obstacle rich and obstacle free terrain, a regime where prior mechanically intelligent limbless systems typically failed due to fixed morphology. Here, Morphing MILR leveraged the behaviors previously described to reach the end of the testing area without any intervention, which is considered a successful trial (Fig.~\ref{fig:transition}). Using the same settings from previous tests for the sidwinding, lateral undulation and rolling, 15 trials were run. The robot needed to progress through the lattice using compliant lateral undulation until a majority of the body had exited the lattice. Rolling then frees the robot entirely from the lattice and changes its heading. Sidewinding continues the traversal outside the lattice until the robot reaches the edge of the testing zone. 

In cases with sufficient clearance, the robot returned to an aligned home configuration before moving on to the next gait. In obstacle rich regions, the robot transitions with high $G$ to prevent mechanical failures and motor overloads. Environment transitions accounted for most of the trial times and were the primary cause behind the failures when the robot tail jams. In 13 of the 15 cases, Morphing MILR successfully escaped the lattice and got to the end of the platform. These experiments demonstrate the central contribution of the platform: the morphology actively reorients without losing the mechanically intelligent benefits of compliance, allowing continuous operation in heterogeneous environments. This transition capability represents a meaningful step toward generalized multi-terrain limbless locomotion, where the robot can adapt morphology on the fly in cramped situations with little sensory awareness.


\section{Conclusion}\label{sec:conclusion}

In this work, we present Morphing MILR, a cable-driven limbless robot that unifies previously separate mechanically intelligent morphologies within a single reconfigurable architecture. Prior systems demonstrated how programmable compliance and cable-driven undulation can simplify control and improve robustness during contact-rich locomotion, but are typically tied to a fixed body configuration and, in many cases, a fixed dominant bending plane. 

Morphing MILR extends this paradigm by pairing bilateral cable-driven bending with body-axis rotation through rolling base joints, allowing the robot to actively remap its bending planes and reorient its body-wave direction in three dimensions. As a result, the platform reproduces multiple  gaits, including lateral undulation, sidewinding, rolling, and twisting while maintaining generalized compliance. Although the present study primarily focused on demonstrating each gait independently, the same modular control structure already enables transitions by reconfiguring joint-plane orientation without requiring the robot to stop its ongoing locomotion sequence.

Moving forward, we will develop continuous gait-transition controllers that leverage this morphology control to smoothly shift among locomotion modes in response to terrain feedback. Beyond the demonstrated gait set, Morphing MILR provides a foundation for synthesizing new three-dimensional motion patterns. Future controllers will exploit geometric descriptions of motion, such as curvature and torsion profiles derived from target trajectories, to map desired three-dimensional paths into coordinated joint commands.In parallel, future work will gather more quantitative data to compare this design against the purpose-built predecessors of \cite{wang2023mechanical,kojouharov2024anisotropic}, determining whether the added complexity yields any change in effectiveness in key metrics such as cost of transport and body lengths per cycle. Together, these results highlight how extending mechanical intelligence from isolated gait-specific designs can be unified into versatile platform, enabling planar, volumetric, and rotational modes of movement. 


\bibliographystyle{IEEEtran}

\bibliography{IROS2026_MorphingMILR}

\begin{thebibliography}{10}
\providecommand{\url}[1]{#1}
\csname url@rmstyle\endcsname
\providecommand{\newblock}{\relax}
\providecommand{\bibinfo}[2]{#2}
\providecommand\BIBentrySTDinterwordspacing{\spaceskip=0pt\relax}
\providecommand\BIBentryALTinterwordstretchfactor{4}
\providecommand\BIBentryALTinterwordspacing{\spaceskip=\fontdimen2\font plus
\BIBentryALTinterwordstretchfactor\fontdimen3\font minus \fontdimen4\font\relax}
\providecommand\BIBforeignlanguage[2]{{%
\expandafter\ifx\csname l@#1\endcsname\relax
\typeout{** WARNING: IEEEtran.bst: No hyphenation pattern has been}%
\typeout{** loaded for the language `#1'. Using the pattern for}%
\typeout{** the default language instead.}%
\else
\language=\csname l@#1\endcsname
\fi
#2}}

\bibitem{hirose1993biologically}
S.~Hirose, ``Biologically inspired robots,'' \emph{Snake-Like Locomotors and Manipulators}, 1993.

\bibitem{hirose2004biologically}
S.~Hirose and M.~Mori, ``Biologically inspired snake-like robots,'' in \emph{2004 IEEE International Conference on Robotics and Biomimetics}.\hskip 1em plus 0.5em minus 0.4em\relax IEEE, 2004, pp. 1--7.

\bibitem{liljeback2012review}
P.~Liljeb{\"a}ck, K.~Y. Pettersen, {\O}.~Stavdahl, and J.~T. Gravdahl, ``A review on modelling, implementation, and control of snake robots,'' \emph{Robotics and Autonomous Systems}, vol.~60, no.~1, pp. 29--40, 2012.

\bibitem{wright2007design}
C.~Wright, A.~Johnson, A.~Peck, Z.~McCord, A.~Naaktgeboren, P.~Gianfortoni, M.~Gonzalez-Rivero, R.~Hatton, and H.~Choset, ``Design of a modular snake robot,'' in \emph{2007 IEEE/RSJ International Conference on Intelligent Robots and Systems}.\hskip 1em plus 0.5em minus 0.4em\relax IEEE, 2007, pp. 2609--2614.

\bibitem{wright2012unified}
C.~G. Wright, A.~D. Buchan, B.~Brown, J.~C. Geist, M.~Schwerin, D.~Rollinson, M.~Tesch, and H.~Choset, ``Design and architecture of the unified modular snake robot,'' in \emph{2012 IEEE International Conference on Robotics and Automation (ICRA)}.\hskip 1em plus 0.5em minus 0.4em\relax IEEE, 2012, pp. 4347--4354.

\bibitem{rollinson2014series}
D.~Rollinson, Y.~Bilgen, B.~Brown, F.~Enner, S.~Ford, C.~Layton, J.~Rembisz, M.~Schwerin, A.~Willig, P.~Velagapudi, and H.~Choset, ``Design and architecture of a series elastic snake robot,'' in \emph{2014 IEEE/RSJ International Conference on Intelligent Robots and Systems (IROS)}.\hskip 1em plus 0.5em minus 0.4em\relax IEEE, 2014, pp. 4630--4636.

\bibitem{crespi2008onlineopt}
A.~Crespi and A.~J. Ijspeert, ``Online optimization of swimming and crawling in an amphibious snake robot,'' \emph{IEEE Transactions on Robotics}, vol.~24, no.~1, pp. 75--87, 2008.

\bibitem{crespi2005amphibotI}
A.~Crespi, A.~Badertscher, A.~Guignard, and A.~J. Ijspeert, ``{AmphiBot I}: an amphibious snake-like robot,'' \emph{Robotics and Autonomous Systems}, vol.~50, no.~4, pp. 163--175, 2005.

\bibitem{transeth2008snake}
A.~A. Transeth, R.~I. Leine, C.~Glocker, K.~Y. Pettersen, and P.~Liljeb{\"a}ck, ``Snake robot obstacle-aided locomotion: Modeling, simulations, and experiments,'' \emph{IEEE Transactions on Robotics}, vol.~24, no.~1, pp. 88--104, 2008.

\bibitem{tesch2009parameterized}
M.~Tesch, K.~Lipkin, I.~Brown, R.~Hatton, A.~Peck, J.~Rembisz, and H.~Choset, ``Parameterized and scripted gaits for modular snake robots,'' \emph{Advanced Robotics}, vol.~23, no.~9, pp. 1131--1158, 2009.

\bibitem{travers2016shapebased}
M.~Travers, J.~Whitman, P.~Schiebel, D.~I. Goldman, and H.~Choset, ``Shape-based compliance in locomotion,'' in \emph{Robotics: Science and Systems (RSS)}, 2016.

\bibitem{saito2002serpentine}
M.~Saito, M.~Fukaya, and T.~Iwasaki, ``Modeling, analysis, and synthesis of serpentine locomotion with a multilink robotic snake,'' \emph{IEEE Control Systems Magazine}, vol.~22, no.~1, pp. 64--81, 2002.

\bibitem{gong2016kinematic}
C.~Gong, M.~J. Travers, H.~C. Astley, L.~Li, J.~R. Mendelson, D.~I. Goldman, and H.~Choset, ``Kinematic gait synthesis for snake robots,'' \emph{The International Journal of Robotics Research}, vol.~35, no. 1-3, pp. 100--113, 2016.

\bibitem{kano2013scaffold}
T.~Kano and A.~Ishiguro, ``Obstacles are beneficial to me! scaffold-based locomotion of a snake-like robot using decentralized control,'' in \emph{2013 IEEE/RSJ International Conference on Intelligent Robots and Systems (IROS)}.\hskip 1em plus 0.5em minus 0.4em\relax IEEE, 2013, pp. 3273--3278.

\bibitem{sanfilippo2017perception}
F.~Sanfilippo, J.~Azpiazu, G.~Marafioti, A.~A. Transeth, {\O}.~Stavdahl, and P.~Liljeb{\"a}ck, ``Perception-driven obstacle-aided locomotion for snake robots: the state of the art, challenges and possibilities,'' \emph{Applied Sciences}, vol.~7, no.~4, p. 336, 2017.

\bibitem{burdick1993sidewinding}
J.~W. Burdick, J.~Radford, and G.~S. Chirikjian, ``A'sidewinding'locomotion gait for hyper-redundant robots,'' in \emph{[1993] Proceedings IEEE International Conference on Robotics and Automation}.\hskip 1em plus 0.5em minus 0.4em\relax IEEE, 1993, pp. 101--106.

\bibitem{marvi2014sidewinding}
H.~Marvi, C.~Gong, N.~Gravish, H.~Astley, M.~Travers, R.~L. Hatton, J.~R. Mendelson~III, H.~Choset, D.~L. Hu, and D.~I. Goldman, ``Sidewinding with minimal slip: Snake and robot ascent of sandy slopes,'' \emph{Science}, vol. 346, no. 6206, pp. 224--229, 2014.

\bibitem{astley2015modulation}
H.~C. Astley, C.~Gong, J.~Dai, M.~Travers, M.~M. Serrano, P.~A. Vela, H.~Choset, J.~R. Mendelson~III, D.~L. Hu, and D.~I. Goldman, ``Modulation of orthogonal body waves enables high maneuverability in sidewinding locomotion,'' \emph{Proceedings of the National Academy of Sciences}, vol. 112, no.~19, pp. 6200--6205, 2015.

\bibitem{zhen2015rolling}
W.~Zhen, C.~Gong, and H.~Choset, ``Modeling rolling gaits of a snake robot,'' in \emph{2015 IEEE International Conference on Robotics and Automation (ICRA)}.\hskip 1em plus 0.5em minus 0.4em\relax IEEE, 2015, pp. 3741--3746.

\bibitem{takemori2018ladder}
T.~Takemori, M.~Tanaka, and F.~Matsuno, ``Ladder climbing with a snake robot,'' in \emph{2018 IEEE/RSJ International Conference on Intelligent Robots and Systems (IROS)}.\hskip 1em plus 0.5em minus 0.4em\relax IEEE, 2018, pp. 8140--8145.

\bibitem{takemori2021hooppassing}
T.~Takemori, M.~Tanaka, and F.~Matsuno, ``Hoop-passing motion for a snake robot to realize motion transition across different environments,'' \emph{IEEE Transactions on Robotics}, vol.~37, no.~5, pp. 1696--1711, 2021.

\bibitem{takemori2022adaptive}
T.~Takemori, M.~Tanaka, and F.~Matsuno, ``Adaptive helical rolling of a snake robot to a straight pipe with irregular cross-sectional shape,'' \emph{IEEE Transactions on Robotics}, vol.~39, no.~1, pp. 437--451, 2022.

\bibitem{wang2021reconstruction}
T.~Wang, B.~Lin, B.~Chong, J.~Whitman, M.~Travers, D.~I. Goldman, G.~Blekherman, and H.~Choset, ``Reconstruction of backbone curves for snake robots,'' \emph{IEEE Robotics and Automation Letters}, vol.~6, no.~2, pp. 3264--3270, 2021.

\bibitem{hanssen2020path}
K.~G. Hanssen, A.~A. Transeth, F.~Sanfilippo, P.~Liljeb{\"a}ck, and {\O}.~Stavdahl, ``Path-planning for perception-driven obstacle-aided snake robot locomotion,'' in \emph{2020 16th International Workshop on Advanced Motion Control (AMC)}.\hskip 1em plus 0.5em minus 0.4em\relax IEEE, 2020, pp. 98--104.

\bibitem{sartoretti2021autonomous}
G.~Sartoretti, T.~Wang, G.~Chuang, Q.~Li, and H.~Choset, ``Autonomous decentralized shape-based navigation for snake robots in dense environments,'' in \emph{2021 IEEE International Conference on Robotics and Automation (ICRA)}.\hskip 1em plus 0.5em minus 0.4em\relax IEEE, 2021, pp. 9276--9282.

\bibitem{ruscelli2018proprioceptive}
F.~Ruscelli, G.~Sartoretti, J.~Nan, Z.~Feng, M.~Travers, and H.~Choset, ``Proprioceptive-inertial autonomous locomotion for articulated robots,'' in \emph{2018 IEEE International Conference on Robotics and Automation (ICRA)}.\hskip 1em plus 0.5em minus 0.4em\relax IEEE, 2018, pp. 3436--3441.

\bibitem{tanaka2015rangesensor}
M.~Tanaka, K.~Kon, and K.~Tanaka, ``Range-sensor-based semiautonomous whole-body collision avoidance of a snake robot,'' \emph{IEEE Transactions on Control Systems Technology}, vol.~23, no.~5, pp. 1927--1934, 2015.

\bibitem{bing2020perceptionaction}
Z.~Bing, C.~Lemke, F.~O. Morin, Z.~Jiang, L.~Cheng, K.~Huang, and A.~Knoll, ``Perception-action coupling target tracking control for a snake robot via reinforcement learning,'' \emph{Frontiers in Neurorobotics}, vol.~14, p. 591128, 2020.

\bibitem{ramesh2022sensnake}
D.~Ramesh, Q.~Fu, and C.~Li, ``Sensnake: A snake robot with contact force sensing for studying locomotion in complex 3-d terrain,'' in \emph{2022 International Conference on Robotics and Automation (ICRA)}.\hskip 1em plus 0.5em minus 0.4em\relax IEEE, 2022, pp. 2068--2075.

\bibitem{sitti2021physical}
M.~Sitti, ``Physical intelligence as a new paradigm,'' \emph{Extreme Mechanics Letters}, vol.~46, p. 101340, 2021.

\bibitem{fu2020robotic}
Q.~Fu and C.~Li, ``Robotic modelling of snake traversing large, smooth obstacles reveals stability benefits of body compliance,'' \emph{Royal Society open science}, vol.~7, no.~2, p. 191192, 2020.

\bibitem{travers2016shape}
M.~J. Travers, J.~Whitman, P.~E. Schiebel, D.~I. Goldman, and H.~Choset, ``Shape-based compliance in locomotion.'' in \emph{Robotics: Science and Systems}, 2016.

\bibitem{serrano2015incorporating}
M.~M. Serrano, A.~H. Chang, G.~Zhang, and P.~A. Vela, ``Incorporating frictional anisotropy in the design of a robotic snake through the exploitation of scales,'' in \emph{2015 IEEE International Conference on Robotics and Automation (ICRA)}.\hskip 1em plus 0.5em minus 0.4em\relax IEEE, 2015, pp. 3729--3734.

\bibitem{wang2023mechanical}
T.~Wang, C.~Pierce, V.~Kojouharov, B.~Chong, K.~Diaz, H.~Lu, and D.~I. Goldman, ``Mechanical intelligence simplifies control in terrestrial limbless locomotion,'' \emph{Science Robotics}, vol.~8, no.~85, p. eadi2243, 2023.

\bibitem{kojouharov2024anisotropic}
V.~Kojouharov, T.~Wang, M.~Fernandez, J.~Maeng, and D.~I. Goldman, ``Anisotropic body compliance facilitates robotic sidewinding in complex environments,'' in \emph{2024 International Conference on Robotics and Automation (ICRA)}.\hskip 1em plus 0.5em minus 0.4em\relax IEEE, 2024.

\bibitem{wang2025aquamilr}
T.~Wang, N.~Mankame, M.~Fernandez, V.~Kojouharov, and D.~I. Goldman, ``Aquamilr: Mechanical intelligence simplifies control of undulatory robots in cluttered fluid environments,'' in \emph{2025 IEEE International Conference on Robotics and Automation (ICRA)}.\hskip 1em plus 0.5em minus 0.4em\relax IEEE, 2025, pp. 14\,671--14\,677.

\bibitem{fernandez2025aquamilrplus}
M.~Fernandez, T.~Wang, G.~Tunnicliffe, D.~Dortilus, P.~Gunnarson, J.~O. Dabiri, and D.~I. Goldman, ``Aquamilr+: Design of an untethered limbless robot for complex aquatic terrain navigation,'' in \emph{2025 IEEE International Conference on Robotics and Automation (ICRA)}, Atlanta, USA, 2025.

\bibitem{wang2026three}
T.~Wang, M.~Fernandez, G.~Tunnicliffe, N.~Cornell, J.~Duong, D.~Dortilus, Z.~J. Xu, P.~Meza, S.~Lublinsky, D.~Parikh, \emph{et~al.}, ``Three-dimensional hydro-cluttered locomotion by an undulatory robot,'' \emph{arXiv preprint arXiv:2606.06829}, 2026.

\bibitem{burdick1994sidewinding}
J.~W. Burdick, J.~Radford, and G.~S. Chirikjian, ``A'sidewinding'locomotion gait for hyper-redundant robots,'' \emph{Advanced Robotics}, vol.~9, no.~3, pp. 195--216, 1994.

\bibitem{chong2021frequency}
B.~Chong, T.~Wang, J.~M. Rieser, B.~Lin, A.~Kaba, G.~Blekherman, H.~Choset, and D.~I. Goldman, ``Frequency modulation of body waves to improve performance of sidewinding robots,'' \emph{The International Journal of Robotics Research}, vol.~40, no. 12-14, pp. 1547--1562, 2021.

\bibitem{schiebel2020robophysical}
P.~E. Schiebel, M.~C. Maisonneuve, K.~Diaz, J.~M. Rieser, and D.~I. Goldman, ``Robophysical modeling of bilaterally activated and soft limbless locomotors,'' in \emph{Biomimetic and Biohybrid Systems: 9th International Conference, Living Machines 2020, Freiburg, Germany, July 28--30, 2020, Proceedings 9}.\hskip 1em plus 0.5em minus 0.4em\relax Springer, 2020, pp. 300--311.

\bibitem{wang2020directional}
T.~Wang, J.~Whitman, M.~Travers, and H.~Choset, ``Directional compliance in obstacle-aided navigation for snake robots,'' in \emph{2020 American Control Conference (ACC)}.\hskip 1em plus 0.5em minus 0.4em\relax IEEE, 2020, pp. 2458--2463.

\end{thebibliography}

\end{document}